# KAN-Matrix: Visualizing Nonlinear Pairwise and Multivariate Contributions for Physical Insight

Luis A. De la Fuente[1], Hernan A. Moreno[1], Laura V. Alvarez[1], and Hoshin V. Gupta[2]

[1]Department of Earth, Environmental and Resource Sciences
University of Texas, El Paso, TX 79902, USA

[2]Department of Hydrology and Atmospheric Sciences
The University of Arizona, Tucson, AZ 85721, USA

Corresponding authors:

Luis A. De la Fuente (ladelafuenteco@utep.edu), ORCID: 0000-0001-6979-0547

Hernan A. Moreno (moreno@utep.edu), ORCID: 0000-0003-0408-6588

Laura V. Alvarez (alvarez@utep.edu), ORCID: 0000-0002-5047-5384

Hoshin V. Gupta (hoshin@arizona.edu), ORCID: 0000-0001-9855-2839

**Key Points**

- Input-output association comprises two components: strength and functional form.
- Pairwise and multivariate association with the target variable provides a deeper understanding of the association.
- The KAN-matrix is a more robust and parsimonious tool compared to traditional approaches.

**Abstract**

Interpreting complex datasets remains a major challenge for scientists, particularly due to high dimensionality and collinearity among variables. We introduce a novel application of Kolmogorov-Arnold Networks (KANs) to enhance interpretability and parsimony beyond what traditional correlation analyses offer.

We present two interpretable, color-coded visualization tools: the *Pairwise KAN Matrix* (PKAN) and the *Multivariate KAN Contribution Matrix* (MKAN). PKAN characterizes nonlinear associations between pairs of variables, while MKAN serves as a nonlinear feature-ranking tool that quantifies the relative contributions of inputs in predicting a target variable.

These tools support pre-processing (e.g., feature selection, redundancy analysis) and post-processing (e.g., model explanation, physical insights) in model development workflows. Through experimental comparisons, we demonstrate that PKAN and MKAN yield more robust and informative results than *Pearson Correlation* and *Mutual Information.* By capturing the strength and functional forms of relationships, these matrices facilitate the discovery of hidden physical patterns and promote domain-informed model development.

Our era is defined by vast data growth and widespread access to machine learning models across scientific and engineering domains. These resources create opportunities for discovery but pose challenges such as high-dimensionality, noise, collinearity, and non-linear relationships that limit interpretability. Large datasets also impose heavy computational demands, restricting analyses and reducing confidence in outcomes. Therefore, tools that efficiently manage complexity while producing interpretable and trustworthy models are needed across disciplines[1].

Scientists and engineers face key challenges during model development: the data must contain sufficient information about the underlying processes[2], be of good quality[3,4], and exhibit causal dependence between inputs and targets[5,6]. When these conditions are not met, models often learn relationships that fail in practice. Tools that support rigorous data examination are essential, enabling researchers to understand the structure, limitations, and causal signals in their data before building models.

In geosciences and other fields, datasets often contain many variables that can provide predictive information about another dependent variable. However, due to lack of clarity of its relative importance, researchers often use *all* inputs variables in ML models, producing overly complex results that are difficult to interpret. Many tools for assessing inter-variable relationships are available, but the most common ones (e.g., *Pearson correlation*, PCA) assume linearity of relationships, while others (e.g., *mutual information, kernel methods, t-SNE*) are either too computationally costly for large datasets or provide only limited insight[7–9].

Many tools reveal only pairwise associations between inputs and a target, overlooking synergistic effects[10] and overestimating correlations when unmeasured factors drive the relationships[11]. While a *Partial Correlation Analysis* can help, it still assumes linearity in the relationships. Therefore, new tools are needed to understand complex data-generating processes (DGPs).

We propose a method that captures both *strength of pairwise associations* (colored line in **Figure 1a**, and colored shading in **Figure 1c**) and also their *functional form* ($\phi$ function in **Figure 1a**, and white/grey lines in **Figure 1c**). We complement the pairwise analysis with a *multivariate association matrix* (MKAN; **Figure 1b** and **Figure 1c**) that considers relationship strengths and functional forms of many input variables, helping to distill simple associations.

The *pairwise association* (PKAN) and *multivariate association* (MKAN) matrices are based on the recently proposed *Kolmogorov-Arnold Network* (KANs)[12] for characterizing model variable associations. The KAN machine learning architecture enhances both interpretability and performance. While its performance relative to traditional neural networks is still under investigation, the KAN approach provides readily interpretable representations[13,14], aligning with our goal of identifying functional relationships and strengths.

Here, we demonstrate the robustness of the PKAN and MKAN tools by comparing them with two common data analysis approaches, and demonstrate their interpretability via two real data experiments. Our goal is to show that these tools can facilitate the development of interpretable, trustworthy models in support of science, stakeholder credibility, and decision-making[16,17].

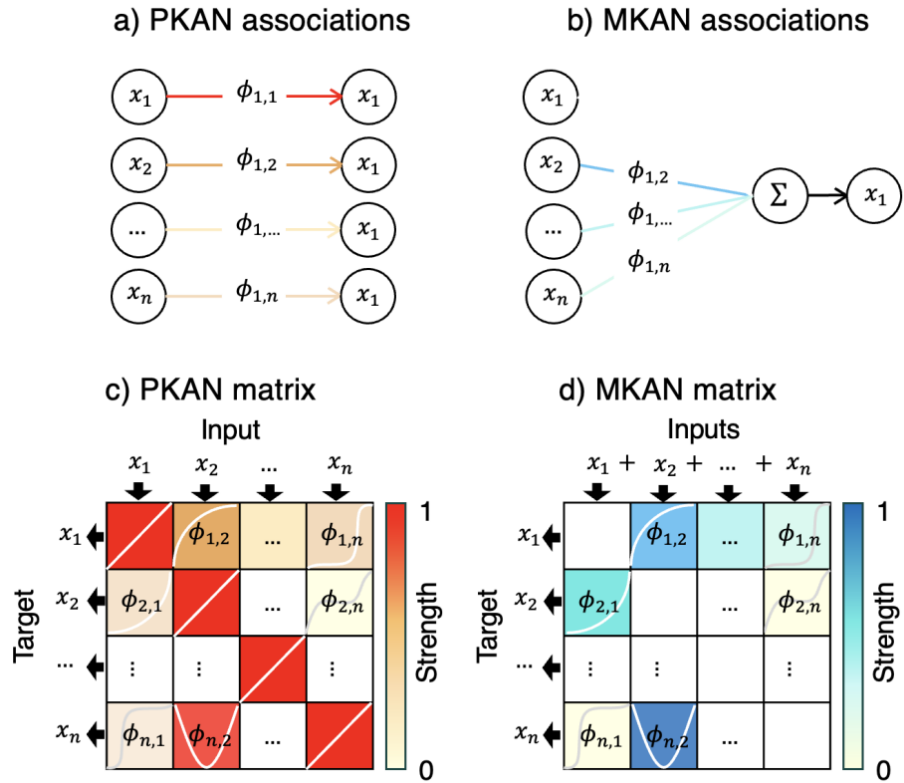

**Fig. 1|** Conceptual representation of *Pairwise KAN (PKAN)* and *Multivariate KAN (MKAN)* matrices. Association strength is indicated by background color intensity (0 = none, 1 = maximum), and the functional form ($\phi$) is shown by the white or grey line in each cell. a) PKAN is derived by analyzing all possible pairwise associations between variables. b) MKAN is obtained when all non-target variables (columns) predict a unique target variable (rows). c) Matrix representation of all the pairwise associations. d) Matrix representation of the multivariate associations. Arrows indicate the direction of information flow.

## Results

**Quantifying the role of functional form in nonlinear associations**

Although data encodes associations between inputs and targets, identifying their existence alone is insufficient; the relationship type, defined by its mathematical or functional form, is essential for model development and for understanding limitations. While existing methods quantify relationship strength, they have limitations. *Pearson Correlation* captures only linear dependencies[18], and *Mutual Information* detects non-linear associations without indicating direction or form. Our first experiment shows how functional form affects strength estimation when using different methods.

We examined data generated by three simple functions: *linear* ($x_1 = x$), *quadratic* ($x_2 = x^2$) and *cubic* ($x_3 = x^3$), when fed the independent variable $x$ drawn from a uniform distribution on [-2,2]. The three variables $x_1$, $x_2$ and $x_3$ were compiled into a "*dataset*", and their associations were investigated. *Pearson Correlation* (**Figure 2a**) fails to quantify the strengths of the quadratic relationships (yellow cells), providing zero linear correlation strength due to its symmetry. *Mutual Information* (**Figure 2b**) did not suffer from this limitation, returning non-zero values for all cells. However, its strength was affected by the relational form, being stronger when it was monotonically non-decreasing.

In contrast, the PKAN matrix (**Figure 2c**) correctly identifies the strength (dark red color) and functional form when using $x_1$ to predict $x_2$ or $x_3$ (*quadratic* or *cubic*; first column), and $x_3$ (treated as independent) to predict $x_1$ or $x_2$ (*linear* or *quadratic*; third column). In the second column, PKAN cannot recognize the strength and functional form when using $x_2$ to predict $x_1$ or $x_3$ due to the injectivity required for such mappings. A quadratic input cannot determine a linear or cubic output through a unique function without splitting the domain into positive or negative regimes.

Interestingly, unlike with *Pearson Correlation* and *Mutual information* methods, the non-symmetric nature of the PKAN matrix (diagonal) suggests non-injective relationships between variables. This adds new information about the associations under investigation.

The MKAN matrix (**Figure 2d**) reports feature contributions (strength) and functional forms when using all column input variables (excluding the target) to predict the target row variable. The first and third rows, predicting $x_1$ from $x_2$ and $x_3$, and $x_3$ from $x_1$ and $x_2$, mirror PKAN results showing that $x_2$ (*quadratic*) provides no useful information for either variable. The second row (predicting $x_2$) uses only $x_1$ (*linear*) despite having $x_3$ (*cubic*) available, reflecting KAN's preference for the *simplest* functional form. This happens because $x_3$ does not provide additional information about $x_2$ beyond what $x_1$ does. Since the relation with $x_3$ would be more complex, $x_3$ is only used if $x_1$ is not available.

PKAN and MKAN matrices provide more robust strength estimates for nonlinear relationships than traditional methods. They reveal when a mapping cannot be approximated due to non-injective associations. The learned functional shape highlights regions of low and high sensitivity between inputs and targets, offering a more detailed characterization of dataset behavior. This insight is valuable to users, as none of the other evaluated methods provide it.

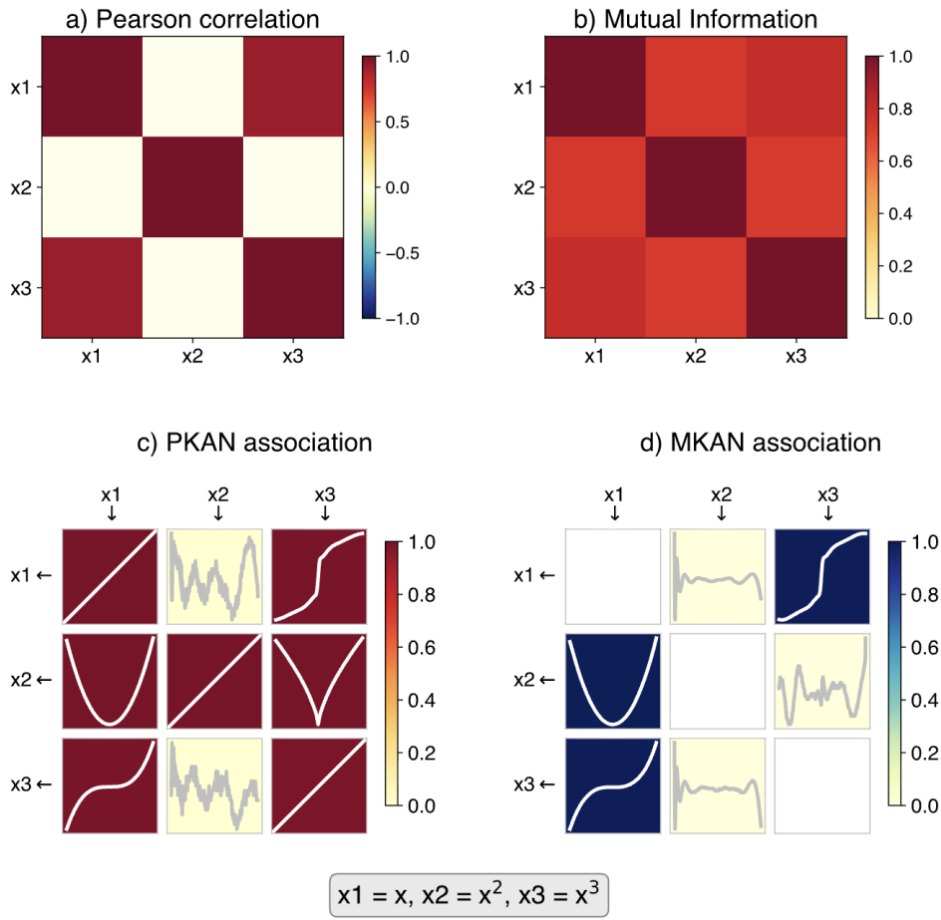

**Fig. 2|** Comparison of methods on non-linear relationships. a) *Pearson correlation* fails to capture quadratic structure. b) *Mutual Information* detects nonlinearity but provides uneven strength estimates. c) PKAN and d) MKAN correctly identify most functional forms, though some quadratic patterns are under-represented due to the non-injective nature of the associated relationships.

**The effect of noise in the associations**

In many real-world settings, measurement error increases with the magnitude of the variable. For example, in hydrology, river discharge is harder to measure during floods, leading to heteroscedastic errors that distort associations. The second experiment, evaluates how different types of noise affect the methods.

We examine the effects of normally-distributed error on a linear relation ($2x$), by adding either homoscedastic noise with a constant standard deviation ($x_2 = 2x + \text{noise}$), or heteroscedastic noise with a variable standard deviation equal to the independent variable ($x_3 = 2x + \text{noise}(x)$). Then, we compiled the variables into a dataset and analyzed their associations (**Figure 3**).

**Figure 3a** shows that noise type (heteroscedastic vs homoscedastic) has a minor effect on *Pearson Correlation.* In contrast, *Mutual information* (**Figure 3b**) degrades substantially because changes in the error distribution alter the underlying data distribution (originally uniform). Instead, the strength captured by the PKAN matrix (**Figure 3c**) is similar to *Pearson Correlation,* indicating a similar effect of noise type.

The functional forms show little effect of noise type when $x_1$ is used as input (first column). Greater deterioration is seen when the independent variable is noisy (second and third columns). When $x_2$ (*linear-homoscedastic*) is the input (second column), the linear pattern remains distinct enough to recover the true functional form. However, when $x_3$ (*linear-heteroscedastic*) is the input variable (third column), the recovered functional forms resemble logarithmic or square-root functions rather than linear, indicating a stronger distortion of the association under heteroscedastic error.

For MKAN (**Figure 3d**, row 1), using $x_2$ and $x_3$ to predict $x_1$ combines both noisy inputs and partially cancels their errors. Rows 2 and 3 show that MKAN consistently selects the noise-free variable ($x_1$), a desirable property. For all methods, but especially *Mutual Information,* noise weakens the strength estimates. Noise also distorts the functional form when present in the input, but this effect can be reduced with multiple noisy inputs. Overall, knowing noise type and magnitude is crucial as they impact association learning.

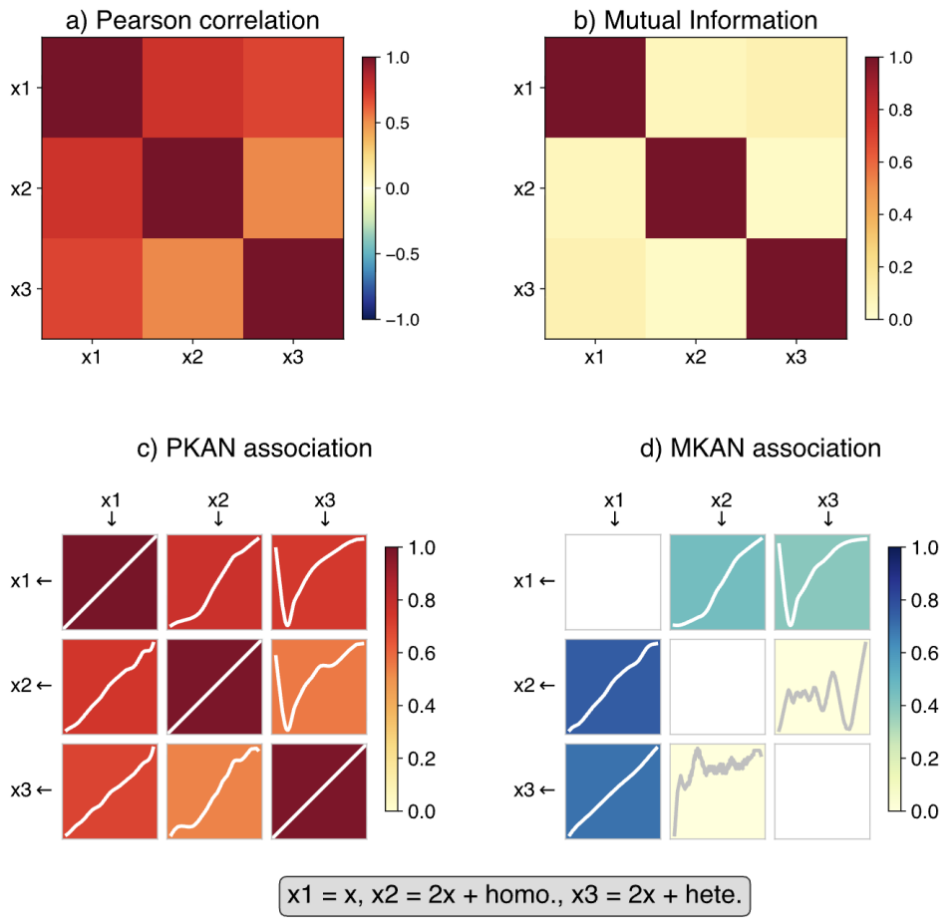

x1 = x, x2 = 2x + homo., x3 = 2x + hete.

**Fig. 3|** Effects of homoscedastic and heteroscedastic data noise on associations. a) *Pearson Correlation* shows moderate degradation. b) *Mutual information* can vanish almost completely under changes in error distribution. c-d) PKAN and MKAN preserve association strength and functional form, although noise in the independent variable distorts the recovered functional form.

**Dynamic associations**

Dynamical processes with lagged dependencies are commonly encountered, but quantification is challenging. To evaluate our method under such conditions, we examined two lagged sinusoidal examples derived from a uniformly distributed variable $x_1$. First, we generated a nonlinear but memoryless transformation ($x_2 = \sin(x_1)$). Second, we introduced memory by lagging $x_1$ by 150 units before applying the sinusoidal transformation ($x_3 = \sin(lagged\ x_1)$). We treated $x_1$, $x_2$, and $x_3$ as different variables and analyzed their association (**Figure 4**).

*Pearson Correlation* fails to capture the strength and lagged effects of the sinusoidal relationships, while *Mutual Information* does better (**Figures 4a** and **4b**). Instead, PKAN accurately characterizes the strengths of the relationships when $x_1$ is used as input (**Figure 4c**; first column) and captured the functional form with minor deviations near the endpoints. The inverse associations (using $x_2$ or $x_3$ to predict $x_1$) were not recovered due to the non-injective nature of the sinusoidal relationship, which requires the superposition of multiple injective functions to represent it. In a full KAN-based model[19,20], increasing the number of hidden-layer nodes between inputs and targets can address this limitation, but to maximize interpretability, our simplified implementation uses only one functional form per cell. MKAN (**Figure 4d**) likewise struggles to predict $x_1$ from the inverse relation (first row) given the non-injective situation, while the second and third rows indicate good performance in predicting $x_2$ and $x_3$ given $x_1$.

KAN matrices outperform traditional approaches even under lagged dependencies. The major limitation arises from the non-injective structure of the inverse relationship, which requires multiple functional forms per cell. Incorporating this capability is left for future extensions.

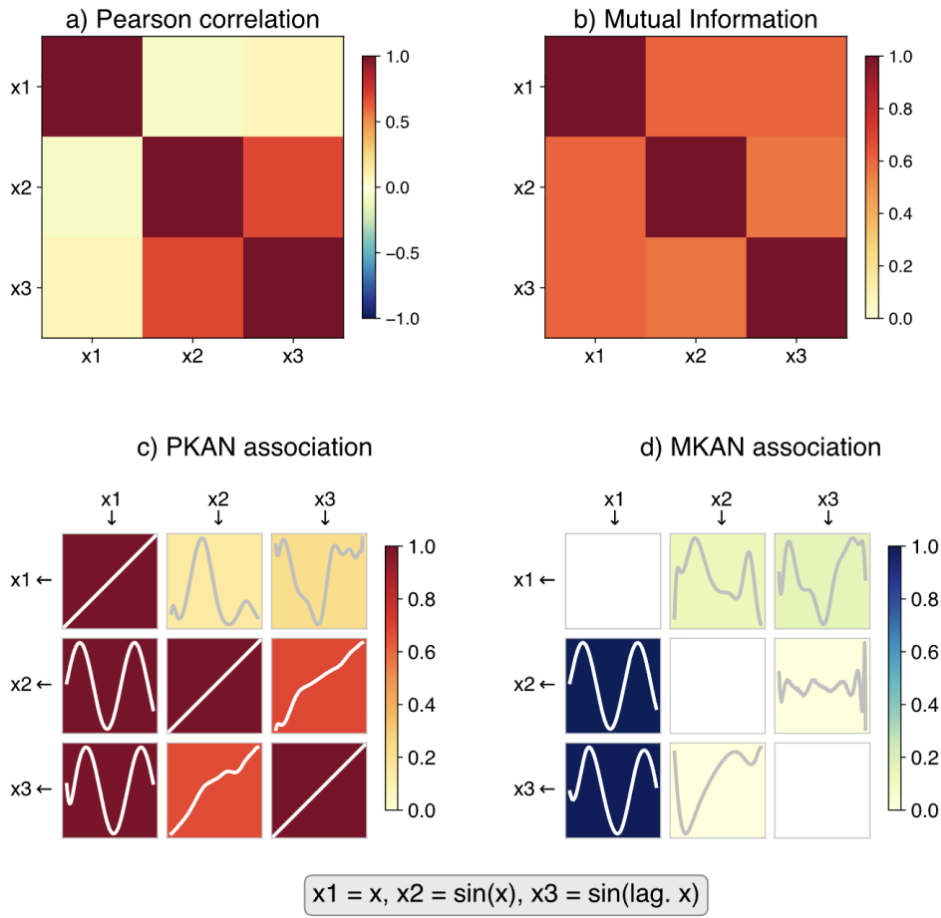

**Fig. 4** | Cyclic and lagged relation experiment. a) *Pearson Correlation* does not reflect the true strength of the sinusoidal or lagged-sinusoidal relationships. b) *Mutual Information* shows only limited improvement. c-d) The KAN matrices accurately capture both strength of association, and functional form of sinusoidal relationships (first column) but fails when the relation is non-injective.

**Multivariate contribution strength and best predictors for hydrologic modeling**

We next use real geoscience data to illustrate the benefits of better estimation and understanding of strength and functional form. Here, we evaluate the multivariate strength contribution (MKAN) of each variable in a dataset and how this can improve model simplicity and performance.

For the first example, we use the CAMELS dataset[21], which provides data on dynamic and static catchment attributes for 671 watersheds across the continental United States. This hydrology-relevant dataset is widely used to build and test models for predicting streamflow and to investigate the interpretability of machine learning models[22–26]. The static attributes span six categories: topography, climate, vegetation, soil, geology, and hydrology (57 attributes).

Given the large number of static attributes, a hydrological analysis typically seeks a minimal and meaningful subset of attributes for streamflow prediction. However, it remains unclear which attributes are most informative for parsimonious model development. Prior studies suggest that only a small subset may be necessary[27], with additional attributes leading to spatial overfitting of data-based models[28]. Here, we use the multivariate MKAN approach to identify attributes that contribute most to predictability, leveraging association strength to guide more parsimonious and robust model design. KAN matrices for other attributes are presented in the supplementary information (S1).

We computed the multivariate contribution (MKAN) for three key streamflow attributes: mean ($q_{mean}$), 5$^{th}$ percentile ($q_5$), and 95$^{th}$ percentile ($q_{95}$). We applied MKAN independently for each attribute as a target, ranking other attributes based on contribution strength. We averaged these strengths across targets to estimate the importance of each attribute for streamflow prediction (multi-target analysis from local MKAN analysis). We used *Pearson Correlation* and *Mutual Information* to generate similar rankings. To assess the practical value of those rankings, we trained a machine learning model to jointly predict the three streamflow attributes using only the top-ranked attributes. We used a *Random Forest* approach[29], which performs well on small dataset[30]. Because percentile values range over several orders of magnitude, we transformed the data logarithmically.

**Table 1** reports the performance of the *Random Forest* model trained with different numbers of top-ranked attributes (from 2 to 14) identified by different ranking methods. For any fixed number of attributes, *MKAN-based* ranking outperformed *Pearson Correlation* and *Mutual Information*. *Pearson-based ranking* (the most common approach) requires 2-6 additional attributes to achieve the same performance as MKAN. This reflects *MKAN*'s ability to capture nonlinear relationships, supporting greater representational parsimony. Most performance gains were obtained using the top 12 MKAN-ranked attributes, while the other methods continued to require additional attributes. *Mutual Information* achieves the second-best performance, likely due to its stronger nonlinearity relative to *Pearson Correlation*.

**Table 1|** Performance of Random Forest models trained using different numbers of top-ranked attributes by different methods (0 is poor and 1 is perfect).

| Ranking approach | R² skill score (Nash-Sutcliffe efficiency in hydrology) | | | | | | |
|---|---|---|---|---|---|---|---|
| | Top 2 | Top 4 | Top 6 | Top 8 | Top 10 | Top 12 | Top 14 |
| Pearson correlation | 0.620 | 0.721 | 0.719 | 0.722 | 0.750 | 0.755 | 0.754 |
| Mutual information | 0.652 | 0.788 | 0.804 | 0.801 | 0.798 | 0.814 | 0.819 |
| MKAN contribution | **0.768** | **0.802** | **0.812** | **0.828** | **0.836** | **0.840** | **0.837** |

**Table 2** reports attribute contributions (strengths) to the three streamflow features used in the MKAN-based ranking. Interestingly, the attributes with the highest contribution strength <u>differ</u> for each streamflow attribute. For example, *frequency of dry days* and *mean slope* contribute to $q_5$ but not to $q_{mean}$ or $q_{95}$, while *mean precipitation* and *forest fraction* contribute more to *qmean* and *q95* but less to *q5*. Moreover, the number of features needed for *qmean* and *q95* is smaller than for *q5* (3 and 5, vs. 12), indicating higher complexity in predicting lower streamflow values. The selection of "*best*" CAMELS attributes depends on the specific target; therefore, any feature selection task should account for variability across the full streamflow distribution.

Table 2| MKAN-derived attribute contributions for predicting streamflow attributes (CAMELS dataset).

| Attribute | $q_5$ | $q_{mean}$ | $q_{95}$ | Average | Ranking |
|---|---|---|---|---|---|
| Mean daily precipitation | 0.109 | 0.623 | 0.526 | 0.419 | 1 |
| Fraction of precipitation falling as snow | 0.080 | 0.182 | 0.144 | 0.135 | 2 |
| Forest fraction | 0.033 | 0.129 | 0.077 | 0.080 | 3 |
| Frequency of dry days (<1 mm/day) | 0.129 | 0.000 | 0.000 | 0.043 | 4 |
| Seasonality and timing of precipitation | 0.000 | 0.001 | 0.104 | 0.035 | 5 |
| Catchment mean slope | 0.097 | 0.000 | 0.000 | 0.032 | 6 |
| Sand fraction | 0.068 | 0.000 | 0.000 | 0.023 | 7 |
| Catchment mean elevation | 0.023 | 0.000 | 0.042 | 0.022 | 8 |
| Difference between the maximum and minimum monthly mean of the leaf area index | 0.058 | 0.000 | 0.000 | 0.019 | 9 |
| Subsurface permeability | 0.055 | 0.000 | 0.000 | 0.018 | 10 |
| Clay fraction | 0.054 | 0.000 | 0.000 | 0.018 | 11 |
| Maximum monthly mean of the green vegetation fraction | 0.050 | 0.000 | 0.000 | 0.017 | 12 |
| Difference between the maximum and minimum monthly mean of the green vegetation fraction | 0.050 | 0.000 | 0.000 | 0.017 | 13 |
| Soil depth | 0.035 | 0.000 | 0.000 | 0.012 | 14 |
| **Total strength contribution** | **0.841** | **0.935** | **0.893** | **0.890** | |

This result illustrates the improved performance and parsimony of the MKAN multivariate contribution analysis for variable selection. It also provides a better understanding of the complexity involved in streamflow prediction and facilitates dimensionality reduction in complex tasks.

**Strength and functional form as a tool for understanding fluid dynamics relationships**

Fluid dynamics are challenging to understand, even knowing the governing partial differential equations[31,32]. Therefore, developing a broad understanding of dominant variables and process relationships is essential. Here, we apply the KAN-based framework in a stagewise fashion, assessing and removing variables to uncover successive levels of feature importance.

The model-generated data analyzed in this experiment are derived from the base-case scenario presented by *España et al.*[33] where the simulated data was produced using a numerical model of the Colorado River reach within Marble Canyon, AZ, USA. In particular, a Large Eddy Simulation (LES) framework resolved the turbulent flow fields[33–35]. Despite the

Navier-Stokes equations being solved, the functional relationships linking the static data ($x, y, z, time$), and the dynamic model outputs, velocity field ($U_x, U_y, U_z$), pressure ($p$), turbulent (eddy) viscosity ($nut$) are not explicitly known. Additionally, lagged values spanning 50 prior time steps were incorporated from the dynamic variables to examine how past system dynamics influence present behavior.

The *PKAN* matrix (**Figure 5a)** indicates high strength for many of the variables. For instance, the dynamic variables ($U_x, U_y, U_z, p, nut$) are associated with coordinates ($x,y$), lagged data, and themselves, due to internal dependency of the variables. However, distinguishing the contribution of each association is infeasible. Instead, the MKAN matrix (**Figure 5b**) shows that most pairwise strengths diminish when lagged simulated data are incorporated, indicating a strong similarity between present and past states. The linear functional form observed between variables and their lagged counterparts (e.g., $U_x = f(U_{x\_lagged})$) suggests persistence effects, indicating the need to remove lagged simulated data to reveal further associations.

Removing the lagged simulated data, as illustrated in **Figure 5c**, reveals that $x$ and $y$ coordinates dominate the contributions to $U_x$ and $U_y$. Since these coordinates lack physical meaning and remain constant, a spatially uniform component, such as the mean, could help characterizing the relationships. Subsequently, the means were subtracted from the velocity, pressure, and turbulent viscosity (creating new variables: $U_x''$, $U_y''$, $U_z''$, $p''$, and $nut''$).

After removing the means (**Figure 5d**), nearly all residual associations vanished except for pressure. The behavior of the delta pressure field ($p''$) is notable, as its strong association with the time variable exceeds the influence of the overall spatial distribution. This temporal interaction is unique to the delta pressure, despite all variables being constrained by the Navier-Stokes equations. These results suggest that the velocity fields are primarily characterized by spatial structure, while pressure exhibits a more pronounced temporal evolution effect. Visualizing and understanding nonlinear strengths via KAN matrices in a sequential fashion provided better insights into controlling associations.

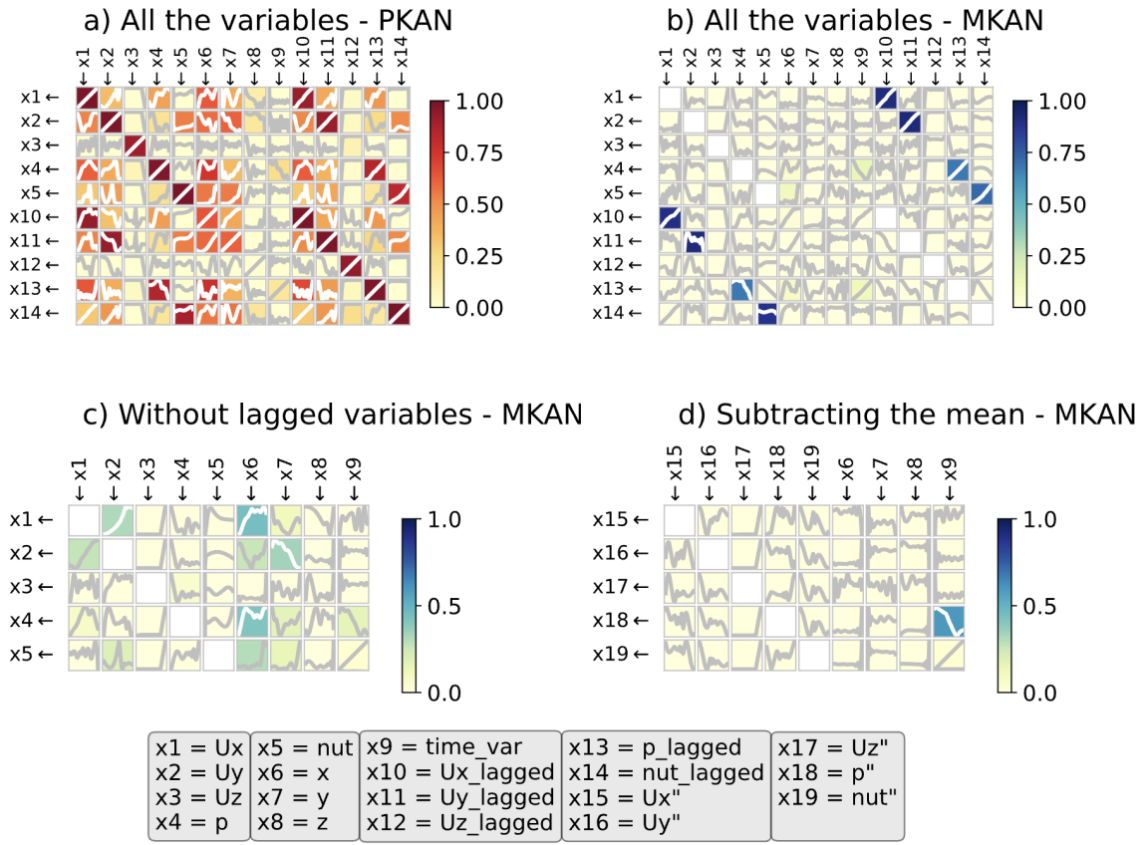

**Fig. 5|** KAN matrices for Computational Fluid Dynamics (non-causal variables *x*, *y*, *z*, and *time_var* were excluded as targets). a) PKAN reveals strong associations between multiple variables. b) MKAN indicates that lagged simulated data dominate the association through a trivial solution. c) MKAN highlights a substantial contribution of the *x* and *y* coordinates to the velocity components. d) MKAN shows that most velocity-related contributions vanish after mean subtraction, reflecting the persistence of an underlying spatial pattern.

## Discussion and Conclusions

Machine learning models can lack interpretability and exhibit "black box" behavior. This motivates the development of tools that make data-model relationships more understandable. However, complex nonlinear, noisy, and multivariate associations can still challenge interpretability during pre- and post-processing. The KAN architecture enables the extraction and visualization of functional forms between variables. We use this characteristic to introduce the pairwise (PKAN) and multivariate (MKAN) matrices, which use simple KAN networks to characterize relational strengths and functional forms between variables.

Traditional methods like *Pearson Correlation* and *Mutual Information* are widely used for assessing dependence, but lack information about relational form, and often struggle with non-linear relationships and noise. Our KAN-based approach is more robust under such conditions. Moreover, it recognizes non-injective relationships through deterioration of interaction strength, a capability not provided by other approaches.

*PKAN* and *MKAN* offer complementary perspectives on the same data. When creating a data-based model, various sources of information are collected to achieve high association with the target (represented by red pixels in the pairwise matrix). However, each interaction may not provide unique information for prediction. *MKAN* finds simple interactions that dominate prediction based on specific inputs. This approach differs from typical machine learning, which often weights all associations (without regularization) to achieve optimal performance.

KAN-based analysis can therefore support feature prioritization in high-dimensional spaces, as demonstrated through the CAMELS attribute optimization experiment. Beyond simplification and enhanced interpretability, pruning improves computational efficiency, making it valuable in large-scale or resource-limited modeling. Accordingly, *KAN* matrices as preprocessing tools guide scientists towards simpler, and more transparent data-based representations. Further, as with the Computational Fluid Dynamics experiment, KAN matrices can facilitate diagnostic post-processing analyses in complex systems where governing mechanisms are often unclear, improving understanding.

Finally, dual characterization of association strength and functional form is critical for understanding any data-generating process. The ability to robustly estimate relational form and strength from data, while providing insights into the direction and structure of information flow, is a common scientific challenge that extends well beyond hydrology and fluid dynamics. Moreover, knowledge of functional form can help to expose hidden knowledge, while providing consistency checks against existing domain knowledge. For these reasons, KAN matrix analysis shows strong potential for improving interpretability in machine learning and deepening scientific understanding across diverse domains.

## Methods

### KAN Architecture

*Liu et al.*[19] introduced the KAN architecture as an extension of the *Kolmogorov-Arnold* theorem (Equation 1[36]), which states that any complex function $y = f(x_1, \ldots x_n)$ can be decomposed into an additive composition of nonlinear univariate transformations $\phi_{i,j}$ of its inputs $x_j$ followed by a further additive composition of nonlinear univariate transformations $\psi_i$ of those components.

$$f(x) = \sum_{i=1}^{2n+1} \psi_i \left[ \sum_{j=1}^{n} \phi_{i,j}(x_j) \right], \tag{1}$$

The theorem does not specify the actual functional natures of the univariate transformation functions $\phi_{i,j}$ and $\psi_i$, only indicating they must exist. *Liu et al.*[19] proposed learning these transformations using spline approximations and expanded the power of the approach by incorporating hierarchical compositional structures resulting in deeper networks. *Liu et al.*[20]

further enhanced the approach by adding a multiplication option and additional features for science discovery.

**The KAN Matrix**

We use a simple KAN architecture that implements only the first level of transformations (i.e., only one $\phi_{i,j}$ function between the input and target layer). This representation treats the inputs as being *informationally mutually independent*, while allowing for non-linear relationships between the inputs and the target variable to exist.

Given that the KAN represents the input-output relationship, the number of inputs used for predicting a target can be arbitrarily defined, allowing evaluation of any possible local association between each input and the target. We visualize this by constructing a matrix of elements, where each element represents the nature of the pairwise association (strength and functional form) between one input and one target (without a $\psi_i$ function). In this case, **Equation 1** reduces to **equation 2**. Plotting the function $\phi_{i,j}$ between each input $x_j$ and target $x_i$, and using the strength of the association provided by the KAN library[37], a unique pairwise association can be characterized. Performing this analysis for each possible combination of input and target generates a matrix of all possible associations (**Figure 1c**). The intersection between each row and column characterizes the *functional nature* of the relation $\phi_{i,j}$, and the intensity of the background color expresses the strength of the association.

$$f_i(x_j) = \phi_{i,j}(x_j), \qquad (2)$$

**Figure 1c** highlights two key properties. First, the diagonal has maximum strength (consistent with perfect association) as the input and target are identical. Second, the matrix need not be symmetric ($\phi_{i,j} \neq \phi_{j,i}$) because the functional relationship may not be injective. In extreme cases, the function $\phi_{i,j}$ may exist but not the function $\phi_{j,i}$.

The pairwise association matrix is simple to visualize and understand, but it does not account for redundant information between variables. A natural extension is to include <u>all</u> relevant variables in the functional representation to predict the selected target. In this case, all column inputs in **Figure 1d**, excluding the target, predict the row target. Therefore, **Equation 1** reduces to **Equation 3**.

$$f_i(x_{i \neq j}, j = 1, \ldots, n) = \sum_{j=1}^{n} \phi_{i,j}(x_j), \qquad (3)$$

In this visualization, functional forms and strengths represent the same as in **Figure 1c**, but are obtained when <u>all</u> inputs predict a target under independence (excluding itself). In this case, the diagonal will not be filled with identity functions because perfect associations are not necessarily achieved. Instead, it represents the relation between the observed and predicted target, predicted using **Equation 3**. An identity function would occur only under a perfect fit. We call this matrix the *Multivariate contribution KAN* (or *MKAN*) because it characterizes the strengths of multiple variables in the learned association with the target.

The strength of the association (indicated by the background color intensity) is defined by the attribute score indicated by *Liu et al.*[20]. They consider the ratio between the standard deviation of edge activation results and the standard deviation of the connected node. This ratio defines the edge's contribution to the node. However, it does not account for the entire network. Therefore, they compute the ratio iteratively, moving from the output layer to the input layer, counting the effective connectivity of edges before and after a specific edge. This method provides a more accurate estimation of feature contribution than the original L1 norm used by *Liu et al.*[19].

In this paper, the *Liu et al.*[20] attribute score is normalized to ensure scores on the range 0-1 for plotting in the KAN matrix. However, that does not consider the predictive strength of the resulting associations. To address this, the attribute score is multiplied by a performance metric that characterizes predictive strength. Two metrics were tested: the *Kling-Gupta Efficiency* (KGE[38]) in its skill-score version [39] and the *$R^2$ skill score* (called *Nash-Sutcliffe Efficiency* in hydrology, NSE[40]). Users can select any metric they prefer, adding interpretability to the strength of association. In our case, both metrics are scaled so that 0.0 corresponds to always using the target mean as the predictor, and 1.0 corresponds to always achieving perfect prediction with one or more inputs. We also use max-min normalization to prevent inputs from having disproportionate importance due to their magnitude in both matrices.

**Details of the Experiments**

Some details regarding the experiment's conceptualization are presented to enhance understanding of the experiment developed.

- In the case of the visualization of *Mutual Information* (*MI*) its normalized version, needed to plot the background color, is not necessarily symmetric. Dividing *Mutual Information* by either the target or input entropies provides insight into the complexity of a variable. For instance, in a Venn diagram, if one set $X$ is contained within another $Y$, the normalized mutual information $NMI(X,Y) = MI(X,Y)/H(Y) = 1$ for the contained set and $NMI(X,Y) = MI(X,Y)/H(X) < 1$ for the containing set, making the normalization non-symmetric. For this paper, we use the average $(H(X) + H(Y))/2$ of the input and target entropies as the normalization, which generates a symmetric normalized mutual information $NMI(X,Y) = 2 \cdot MI(X,Y)/(H(X) + H(Y))$. Non-symmetric normalization can be used instead, but those results do not add value to this discussion, and therefore, they are included in the supplementary information.

- The mathematical expressions used to generate the synthetic data in these experiments are reported in **Table 3**.

**Table 3|** Expressions used to simulate different kinds of behaviors in synthetic data (note that a random sample with a normal distribution is indicated as *Random(μ, σ)* where $\mu$ represents the mean and $\sigma$ represents the standard deviation).

| Behavior experiment | Expression |
|---|---|
| Nonlinear relation | x = (-2,2) Uniform distribution<br>y = $x^2$ + Random(0, 0.1)<br>z = $x^3$ + Random (0, 0.1) |
| Heteroscedasticity | x = (0,10) Uniform distribution<br>y = 2x + Random (0, x)<br>z = 2x + Random (0, 5.0) |
| Lagged relation | x = (0,10) Uniform distribution<br>y = sin(x) + Random (0,0.1)<br>z = sin(shift = 150) + Random (0,0.1) |

- The data split used with the Random Forest models was 80% for training and 20% for testing. The model architecture was kept fixed across ranking approaches, with a fixed seed used for randomization. The hyperparameters represent standard values used with this architecture, and optimization of these hyperparameters was not pursued, as the purpose was to compare methods rather than achieve the best performance. The hyperparameters are presented in **Table 4**.

**Table 4|** Hyperparameter used with Random Forest models.

| Hyperparameter | Value |
|---|---|
| Number of estimators | 100 |
| Minimum samples splitting | 4 |
| Minimum samples leaf | 3 |
| Maximum feature | 0.7 |
| Maximum samples | 0.7 |
| Bootstrapping | True |
| Random seed | 42 |

## Data Availability

The Jupyter Notebooks used to generate figures and tables are freely available on GitHub (https://github.com/ldelafue/KAN_matrix). The Python code can be cloned from the same repository.

## Acknowledgements

This material is based upon work supported by the National Science Foundation (NSF) under Grant No. 2239550 CAREER award, the Army Research Office Grant No. W911NF-24-1-0296 and the National Oceanic and Atmospheric Administration (NOAA) Cooperative Science Center Educational Partnership Program with Minority Serving Institutions (MSI) award

## Author contributions

L.D. designed the study, coded the tool, developed the analysis, and wrote the initial draft of the paper. H.G., H.M., and L.A. provided foundational ideas and actively participated in the editing and revisions of the paper. L.A. provided the turbulence model output simulations and L.A. and H.M. managed the funding documentation for the research.

# Supplementary information

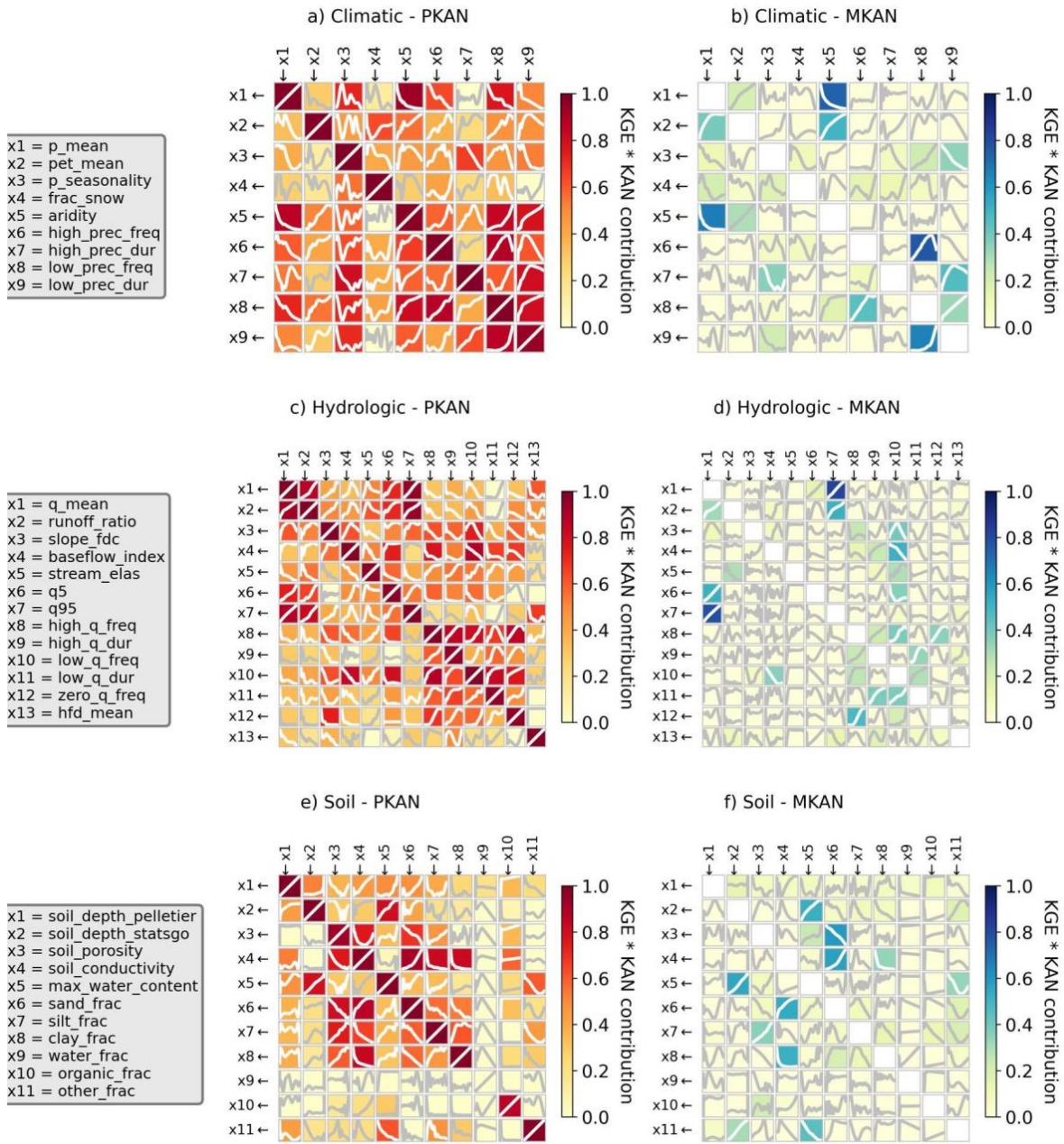

**Fig. S1| P**KAN and MKAN for each of the CAMELS attribute categories. PKAN (left column) shows many strong pairwise associations, suggesting substantial redundancy. MKAN (right column) reveals that only a small subset of attributes contributes meaningfully in a multivariate setting, highlighting the most informative and non-redundant predictors.

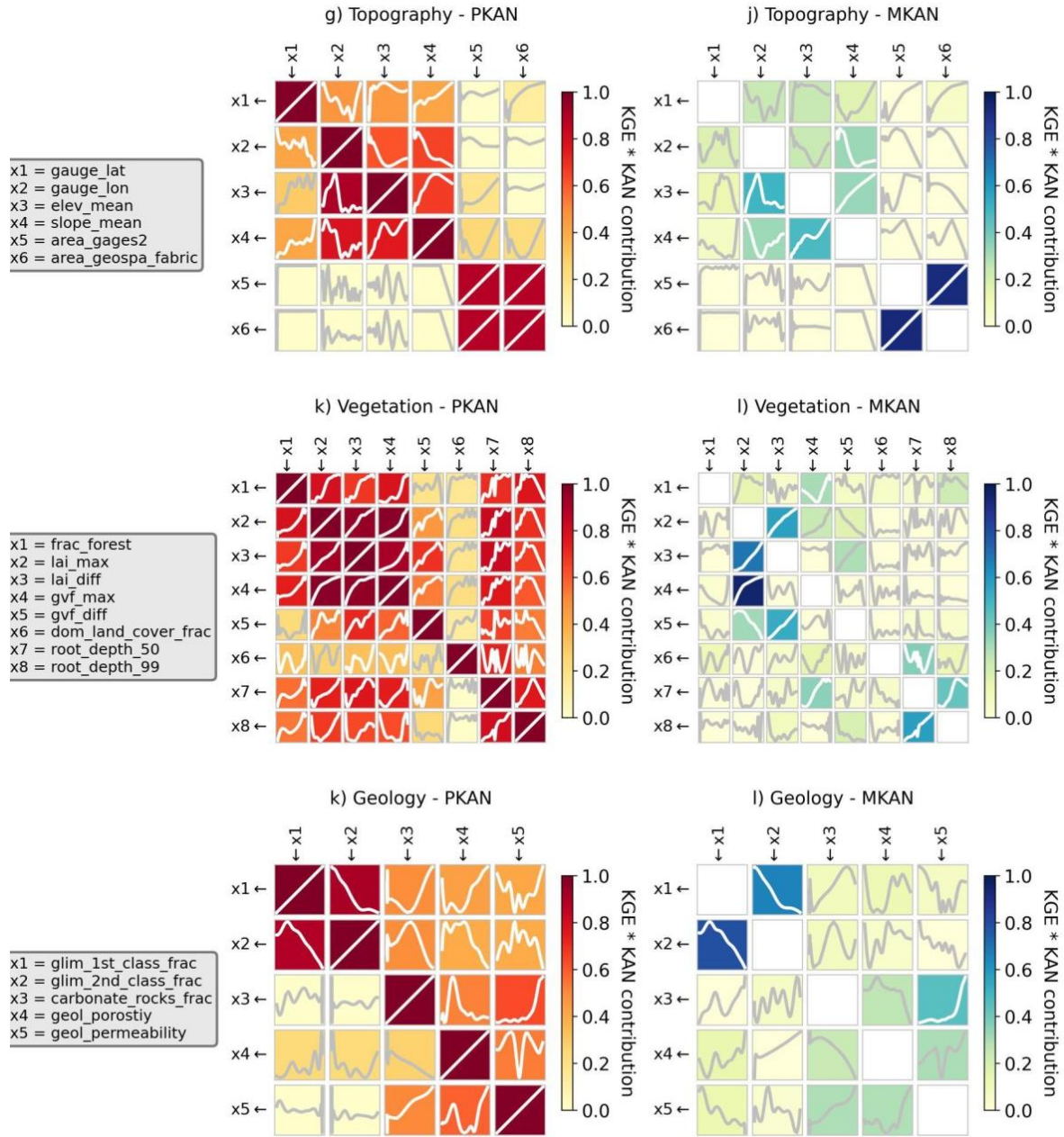

**Fig. S1 (continuation)|** **P**KAN and MKAN for each of the CAMELS attribute categories. PKAN (left column) shows many strong pairwise associations, suggesting substantial redundancy. MKAN (right column) reveals that only a small subset of attributes contributes meaningfully in a multivariate setting, highlighting the most informative and non-redundant predictors.